%% file: main.tex
\documentclass[10pt,twocolumn,letterpaper]{article}
\usepackage[pagenumbers]{cvpr}
\input{preamble}

\definecolor{cvprblue}{rgb}{0.21,0.49,0.74}
\usepackage[pagebackref,breaklinks,colorlinks,citecolor=cvprblue]{hyperref}

\usepackage[capitalize]{cleveref}
\crefname{section}{Sec.}{Secs.}
\Crefname{section}{Section}{Sections}
\Crefname{table}{Table}{Tables}
\crefname{table}{Tab.}{Tabs.}

\begin{document}

\title{\vspace{-1mm}\ours: Video-conditioned Text Representations for Activity Recognition\vspace{-1mm}}

\author{Kumara Kahatapitiya\textsuperscript{1}\thanks{Work done as a student researcher at Google.}\hspace{1mm}\hspace{2mm} Anurag Arnab\textsuperscript{2}\hspace{2mm} Arsha Nagrani\textsuperscript{2}\hspace{2mm} Michael S. Ryoo\textsuperscript{1,2} \vspace{1mm}\\ 
\textsuperscript{1}Stony Brook University \hspace{2mm} \textsuperscript{2}Google Research\\
{\tt\small kkahatapitiy@cs.stonybrook.edu}
}

\maketitle

\input{sections/1_abstract}
\input{sections/2_introduction}

\input{sections/3_related_work}
\input{sections/4_method}
\input{sections/5_experiments}
\input{sections/6_conclusion}
\input{sections/7_appendix}

{\small
\bibliographystyle{ieeenat_fullname}
\bibliography{egbib}
}

\end{document}

%% file: preamble.tex
\usepackage{times}
\usepackage{epsfig}
\usepackage{graphicx}
\usepackage{amsmath}
\usepackage{amssymb}
\usepackage{balance}
\usepackage{nicefrac}
\usepackage{microtype}
\usepackage{pifont}
\usepackage{multirow}
\usepackage{verbatim}
\usepackage{color}
\usepackage{float}
\usepackage{enumitem}
\usepackage{booktabs}
\usepackage{tabulary,multirow,overpic,xcolor}
\usepackage{pifont}
\usepackage{epstopdf}
\usepackage{algorithm}
\usepackage{algorithmicx}
\usepackage{algpseudocode}
\usepackage{enumerate}
\usepackage{bm}
\usepackage{t1enc}
\usepackage{colortbl}
\usepackage{soul}
\usepackage{arydshln}
\usepackage{tabu}
\usepackage[accsupp]{axessibility}

\def\eg{\emph{e.g}\onedot} 

\def\ie{\emph{i.e}\onedot, }

\definecolor{pos}{RGB}{0,153,0}
\definecolor{neg}{RGB}{0,0,0}
\definecolor{smpos}{RGB}{0,0,0}
\definecolor{row}{RGB}{235, 245, 251}
\definecolor{down}{RGB}{153, 163, 164}

\newcommand{\cmark}{\ding{51}}
\newcommand{\xmark}{\ding{55}}

\newcommand{\fref}[1]{Fig.~\ref{#1}}
\newcommand{\tref}[1]{Table~\ref{#1}}

\newcommand{\ours}{\texttt{VicTR}}
\newcommand{\Tau}{\mathcal{T}}

\newlength\savewidth
\newcommand{\tablestyle}[2]{\setlength{\tabcolsep}{#1}\renewcommand{\arraystretch}{#2}\centering\footnotesize}

\makeatletter\renewcommand\paragraph{\@startsection{paragraph}{4}{\z@} {.5em \@plus1ex \@minus.2ex}{-.5em}{\normalfont\normalsize\bfseries}}\makeatother

%% file: sections/1_abstract.tex
\begin{abstract}
   \vspace{-2mm} Vision-Language models (VLMs) have excelled in the image-domain--- especially in zero-shot settings--- thanks to the availability of vast pretraining data (i.e., paired image-text samples). However for videos, such paired data is not as abundant. Therefore, video-VLMs are usually designed by adapting pretrained image-VLMs to the video-domain, instead of training from scratch. All such recipes rely on augmenting visual embeddings with temporal information (i.e., image $\rightarrow$ video), often keeping text embeddings unchanged or even being discarded. In this paper, we argue the contrary, that better video-VLMs can be designed by focusing more on augmenting text, rather than visual information. More specifically, we introduce Video-conditioned Text Representations~(\ours): a form of text embeddings optimized w.r.t.~visual embeddings, creating a more-flexible contrastive latent space. Our model can further make use of freely-available semantic information, in the form of visually-grounded auxiliary text (e.g. object or scene information). We evaluate our model on few-shot, zero-shot (HMDB-51, UCF-101), short-form (Kinetics-400) and long-form (Charades) activity recognition benchmarks, showing strong performance among video-VLMs. 
   \vspace{-6mm}
\end{abstract}

%% file: sections/2_introduction.tex
\section{Introduction}
\label{sec:intro}

Video understanding poses significant challenges, often adding to the complications in image domain such as model complexity and annotation costs. The additional temporal dimension and different modalities of data introduce useful cues, but also can be redundant, raising interesting questions about trade-offs. Activity Recognition (\ie classification) in particular--- as the prominent task in video understanding--- has long been explored by the community in these research directions. Whether it is efficient architecture variants ranging from CNNs \cite{lin2019tsm, feichtenhofer2020x3d, ryoo2019assemblenet} to Transformers \cite{arnab2021vivit, bertasius2021timesformer, fan2021mvit}, training schemes from fully-supervised \cite{carreira2017i3d, feichtenhofer2019slowfast} to self-supervised \cite{qian2021spatiotemporal, feichtenhofer2021large, recasens2021brave} or data regimes from unimodal \cite{xie2017s3d, tran2018r2p1d} to multimodal \cite{han2020self, nagrani2021bottleneck}, the progress has been steady and exciting. More recently, with the availability of internet-scale paired image-text data, the direction of vision-language models (VLMs) \cite{radford2021clip, jia2021align} have emerged dominant, achieving strong generalization across numerous benchmarks. However, the progress of VLMs in the video domain is yet to be caught-up to its full potential.

\begin{figure}[t!]
    \centering
    \includegraphics[width=0.8\linewidth]{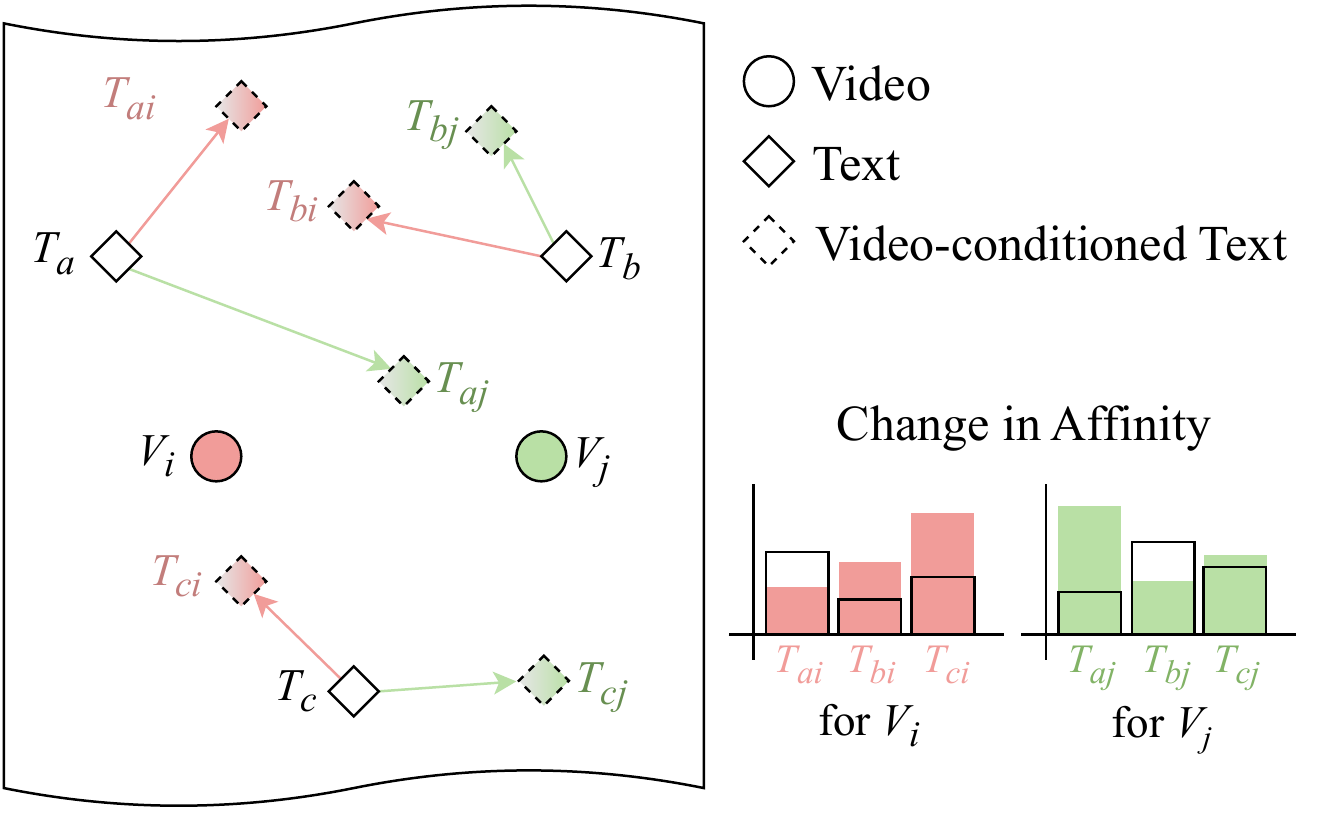}
    \caption{\textbf{Video-conditioned Text Representations:} Pretrained image-VLMs can generate reasonable visual embeddings for videos (\eg by temporally-pooling frame embeddings), together with paired text embeddings. However, usually, these text embeddings are not dependent on visual information--- meaning, they are common for every video. Such representations lack the flexibility to align properly in a shared vision-language latent space, when optimized based on a contrastive similarity (\ie \textit{Affinity}) w.r.t.~all videos. However, with \textit{Video-conditioned Text} representations that specialize uniquely for each video, we grant more freedom for text embeddings to move in the latent space, and adapt to different scenarios (\eg more-challenging recognition tasks).
    }
    \vspace{-4mm}
    \label{fig:concept}
\end{figure}
        
\begin{figure}[t!]
    \centering
    \includegraphics[width=0.9\linewidth]{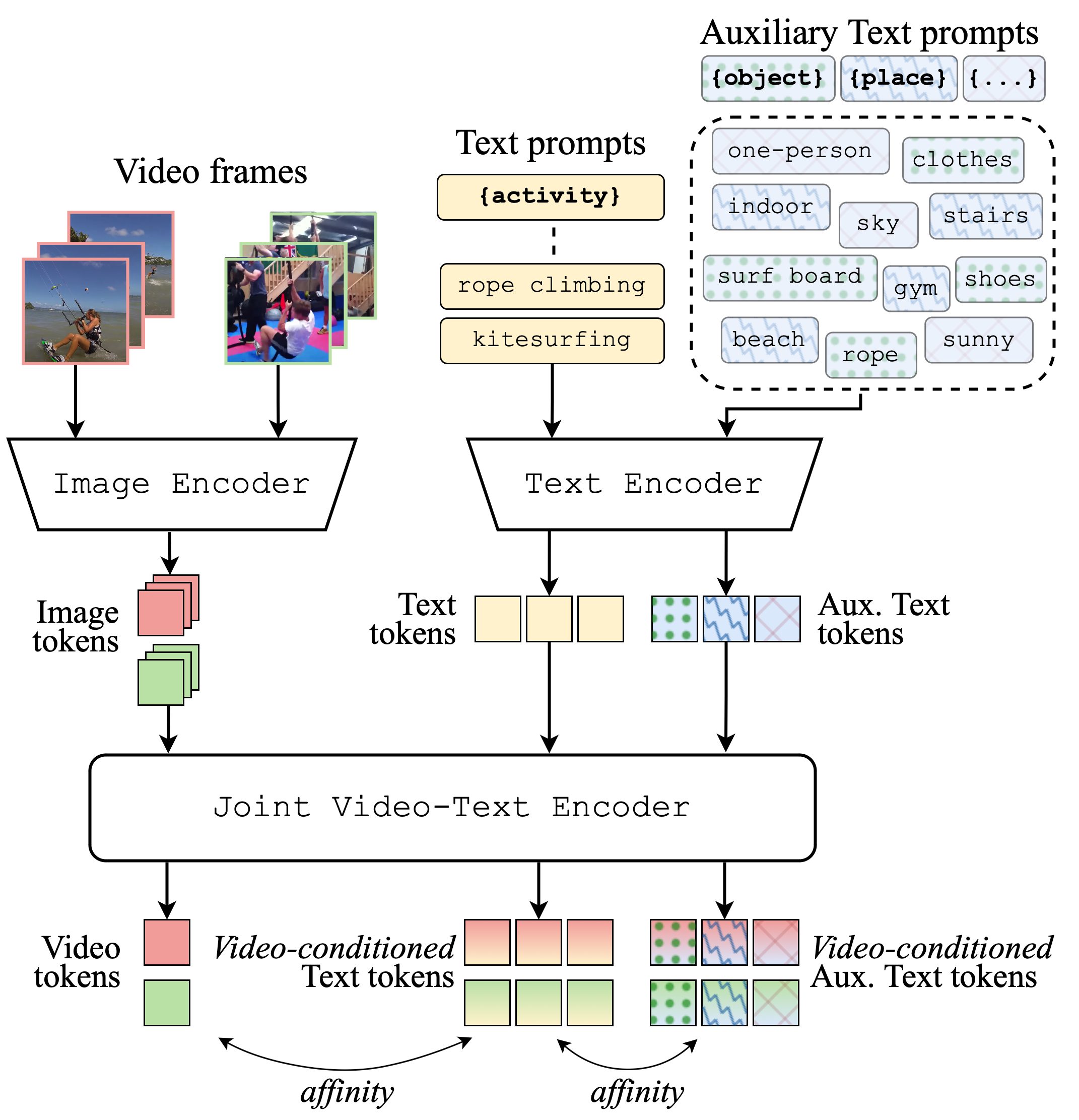}
    \vspace{-2mm}
    \caption{\textbf{Overview of ~\ours:} First, we extract image (\ie frame) and text tokens using a pretrained image-VLM. Next, such tokens go through a joint video-text encoder, generating video tokens and \textit{video-conditioned} text tokens, based on which, we compute affinity-based logits for classification. Optionally, any semantic concept (given as auxiliary text) can also be processed similarly, to help guide the classifier. This is motivated based on the co-occurrence of semantics (\eg \texttt{rope}, \texttt{gym}, \texttt{one-person}) and categories-of-interest, \ie activity classes in our setting (\eg \texttt{rope climbing}). Here, the color change of text tokens represents the idea of video-conditioning.}
    \label{fig:overview}
    \vspace{-4mm}
\end{figure}

Following the seminal VLMs such as CLIP \cite{radford2021clip} and ALIGN \cite{jia2021align}, there have been significant strides in tasks such as image classification \cite{yuan2021florence, zhai2022lit, yan2022videococa}, open-vocabulary object detection \cite{gu2021vild, minderer2022simple}, text-to-image retrieval \cite{singh2022flava, yao2021filip} and robot manipulation \cite{jiang2022vima, zeng2022socratic}. Such models are usually pretrained on paired image-text data based on a contrastive learning framework. The idea is to have two separate backbones--- an Image Encoder and a Text Encoder, that generate embeddings in a joint latent space. To optimize this space, the corresponding pairs of embeddings are drawn closer, by increasing their similarity (\ie \textit{Affinity}). The key advantage of such models is that, at inference, any semantic concept (given as a text input) can be embedded in the same space, giving intriguing zero-shot or few-shot transfer capabilities \cite{zeng2022socratic, alayrac2022flamingo}. For instance, CLIP \cite{radford2021clip} excels at classifying unseen attribute categories (\eg objects, scenes), or even counting such occurrences \cite{zeng2022socratic}. However, these VLMs do not perform well in tasks that require specialized knowledge, such as localizing (\eg detection/segmentation) or temporal reasoning (\eg activity recognition), at least not out-of-the-box, as their training objective has not seen any location or temporal cues. Yet, with task-specific finetuning, such models can readily be adapted to specialized domains \cite{gu2021vild, ma2022xclip}.

In the video domain, training VLMs from scratch may show a limited success \cite{xu2021videoclip}--- while also being expensive--- due to the lack of paired data at scale. As a compromise, the common practice is to adapt pretrained image-VLMs to video, by introducing temporal information. Such methods either insert temporal modules within the image backbone itself to have cross-frame interactions \cite{ma2022xclip}, or use a post-processing video head on-top of the image backbone \cite{luo2022clip4clip, wang2021actionclip, bain2022cliphitchhiker, lin2022evl}. In both cases, image embeddings are enhanced as video embeddings. However, the use of text embeddings varies among different approaches. Text may either be discarded \cite{lin2022evl}, kept frozen \cite{luo2022clip4clip, wang2021actionclip}, used as conditioning \cite{bain2022cliphitchhiker} (to further enhance video embeddings), or fully-updated jointly with video \cite{ma2022xclip}. More often than not, the main focus is on visual embeddings (\ie converting image $\rightarrow$ video), and the impact of updating text has been limited.

Nevertheless, video models benefit from semantic information \cite{ji2020action, zeng2022socratic, wang2022language}. In fact, certain attributes (\eg objects, scene or human subjects) are directly tied with specific activities, and can simplify their recognition. For instance, the presence of attributes such as [\texttt{rope}, \texttt{gym,} \texttt{one-person}] can narrow down the potential activity to \texttt{battling ropes} or \texttt{rope climbing}. VLMs are especially suited to take advantage of such semantics. Any concept represented as text can be visually-grounded based on paired embeddings (in zero-shot), to extract relevant attributes for a given input that benefit recognition tasks. Such visually-grounded semantics are cheap in-terms of both annotation and compute costs, yet highly-useful.

Motivated by the above we propose \ours, focusing on adapting text information to the video domain. More specifically, we generate \textit{Video-conditioned Text} embeddings (see \fref{fig:concept}), while jointly-training both textual and visual features generated by an image-VLM. 
By finetuning text embeddings, we observe significant gains in our framework, compared to just finetuning visual embeddings (similar to the observations in \cite{zhai2022lit}). We can also make use of freely-available auxiliary semantic information, represented in the form of visually-grounded text embeddings. 
\fref{fig:overview} shows an overview of the proposed architecture. Our video-conditioned text embeddings are unique to each video, allowing more-flexibility to move in the latent space and generalize to complex downstream tasks. Optionally, our video-conditioned auxiliary text can further help optimize this latent space. We evaluate \ours~on few-shot, zero-shot, short-form and long-form activity recognition, validating its strong generalization capabilities among video-VLMs.
\vspace{-2mm}

%% file: sections/3_related_work.tex
\section{Related Work}
\label{sec:related_work}

\paragraph{Video understanding} is about reasoning based on spatio-temporal inputs. Compared to image inputs, videos bring additional useful cues such as motion or multiple modalities (\eg audio) into play, but also any associated complications such as increased compute requirements and redundancy in data. Convolutional networks (CNNs) \cite{carreira2017i3d, xie2017s3d, tran2018r2p1d, wang2018nonlocal} and Recurrent models \cite{escorcia2016daps, yeung2018every} have been the state-of-the-art in video modeling, prior to the rise of Transformers \cite{arnab2021vivit, bertasius2021timesformer, liu2022videoswin, ryoo2021tokenlearner}. Multi-stream models \cite{carreira2017i3d, feichtenhofer2019slowfast} that make use of different spatio-temporal views \cite{feichtenhofer2019slowfast, recasens2021brave} or modalities (\eg optical-flow \cite{carreira2017i3d, han2020self}, audio \cite{nagrani2021bottleneck, huang2022mavil, recasens2023zorro}) have emerged, tackling benchmark tasks such as activity recognition \cite{kay2017kinetics, kuehne2011hmdb}, localization \cite{sigurdsson2016hollywood, gu2018ava, yeung2018every} or text-to-video retrieval \cite{xu2016msrvtt}. To handle longer video inputs, models have focused on efficient temporal modeling \cite{piergiovanni2018superevents, piergiovanni2019tgm, kahatapitiya2021coarsefine}, or memory mechanisms \cite{wu2019long, wu2022memvit, ryoo2022ttm}. While Neural Architecture Search (NAS) has enabled efficient model designs \cite{feichtenhofer2020x3d, ryoo2019assemblenet, ryoo2020assemblenet++}, self-supervised methods \cite{recasens2021brave, han2020self, qian2021spatiotemporal, feichtenhofer2021large} have alleviated the high demand for annotated data. More recently, language-supervision has been of interest for video understanding due to the strong generalization capabilities shown in the image domain.
\vspace{-2mm}

\paragraph{Vision-Language Models (VLMs)} are usually trained on internet-scale paired visual-language (\eg image-text) data. Seminal work such as CLIP \cite{radford2021clip} and ALIGN \cite{jia2021align} have shed the light on the capabilities of such models, especially for zero-shot transfer. Since then, VLM literature has flourished, with applications in open-vocabulary object detection \cite{gu2021vild, minderer2022simple}, open-set classification \cite{qian2022multimodal}, retrieval \cite{singh2022flava, yao2021filip, bain2021frozen}, captioning \cite{yang2023vid2seq}, segmentation \cite{xu2022groupvit, ranasinghe2022perceptual}, robot manipulation \cite{zeng2022socratic, jiang2022vima, karamcheti2023language} and many other domains. Although VLMs are generally trained on image-text data, there are intuitive variants which are trained either only on images \cite{tschannen2022image} or only on text \cite{nukrai2022text}. The commonly-used similarity-based objective of VLMs has also been repurposed to specialized domains, through prompt learning \cite{zhou2022learning} or engineering \cite{gu2021vild, paiss2023teaching}. The text encoder of VLMs can be a powerful mapping from semantic concepts to latent embeddings \cite{menon2022visual}. Many foundation models \cite{yuan2021florence, alayrac2022flamingo, yu2022coca} follow similar design principles as VLMs, thriving in zero-shot \cite{guo2022calip} or few-shot \cite{zhou2022learning} settings. Recent work combining Large Language Models (LLMs) with VLMs show how language can act as a communication-medium between models \cite{zeng2022socratic, zhao2022learning, wang2022language}. In \cite{menon2022visual}, authors use an LLM to represent object classes as a set of its semantic attributes, to learn a better classifier. 

As for video-VLMs, they are either trained from scratch on video-text data \cite{xu2021videoclip, yang2023vid2seq}, or more-often than not, finetuned initializing from a pretrained image-VLM \cite{cheng2022vindlu, yan2022videococa, lin2022egocentric}. Some are even trained on both image and video data paired with text \cite{bain2021frozen}. The success of VLMs in the image domain has fueled similar research directions in the video domain.
\vspace{-2mm}

\paragraph{Adapting image-text models to video} is a common practice when designing video-VLMs. A general and effective recipe for such adaptation is proposed in \cite{cheng2022vindlu}. It consists of temporal modeling, multi-modal fusion, auxiliary training objectives, and both image/video data at scale. All others usually make use of a subset of these concepts. CLIP-ViP \cite{xue2022clipvip} is trained with different sources of data and multiple cross-modal training objectives. VideoCoCa \cite{yan2022videococa} extends CoCa \cite{yu2022coca} with attention-pooled frame embeddings, which are used to decode text captions in a generative framework. MOV \cite{qian2022multimodal} is trained with additional audio/flow encoders through cross-modal attention, keeping image-text encoders frozen. Video-specific prompts can also be learned with such frozen encoders \cite{ju2022prompting}. Vi-Fi \cite{rasheed2022fine} shows that simply finetuning CLIP image-text encoders without any specialized modules can generate video representations efficiently.

Apart from the above, there exists a body of prior work that closely-relates to \ours. ActionCLIP \cite{wang2021actionclip} upgrades its CLIP image-encoder with (1) parameter-free temporal layers (TSM \cite{lin2019tsm}) within the backbone, and (2) a temporal transformer head, while keeping the text-encoder fixed. Similarly, CLIP4clip \cite{luo2022clip4clip} just uses a temporal transformer head to update visual embeddings. CLIPHitchhiker's \cite{bain2022cliphitchhiker} generates \textit{text-conditioned video} embeddings by temporally-pooling frame embeddings, conditioned on each text query. In this case, a given video generates multiple different visual embeddings, one per each text embedding. EVL \cite{lin2022evl} completely discards text. It acts as an initialization for a visual-only backbone, consisting of CLIP image encoder and a temporal, class-conditioned decoder. X-CLIP \cite{ma2022xclip} introduces trainable temporal layers within its backbone image encoder, and generates video-specific text prompts. Meaning, it finetunes both encoders similar to ours. However, it does not allow interaction among text embeddings, nor with fine-grained visual information (but only, with temporally-aggregated information). Hence, it shows limited gains from adapting text to video domain. In contrast, our \textit{video-conditioned text} embeddings that are unique for each video, interacts with both fine-grained visual embeddings and other text embeddings, to enable a better contrastive framework, and in-turn, a more-flexible alignment in the latent space.

%% file: sections/4_method.tex
\section{Background: image-VLMs to video}
\label{sec:background}

\begin{figure*}[ht!]
    \centering
    \includegraphics[width=0.9\linewidth]{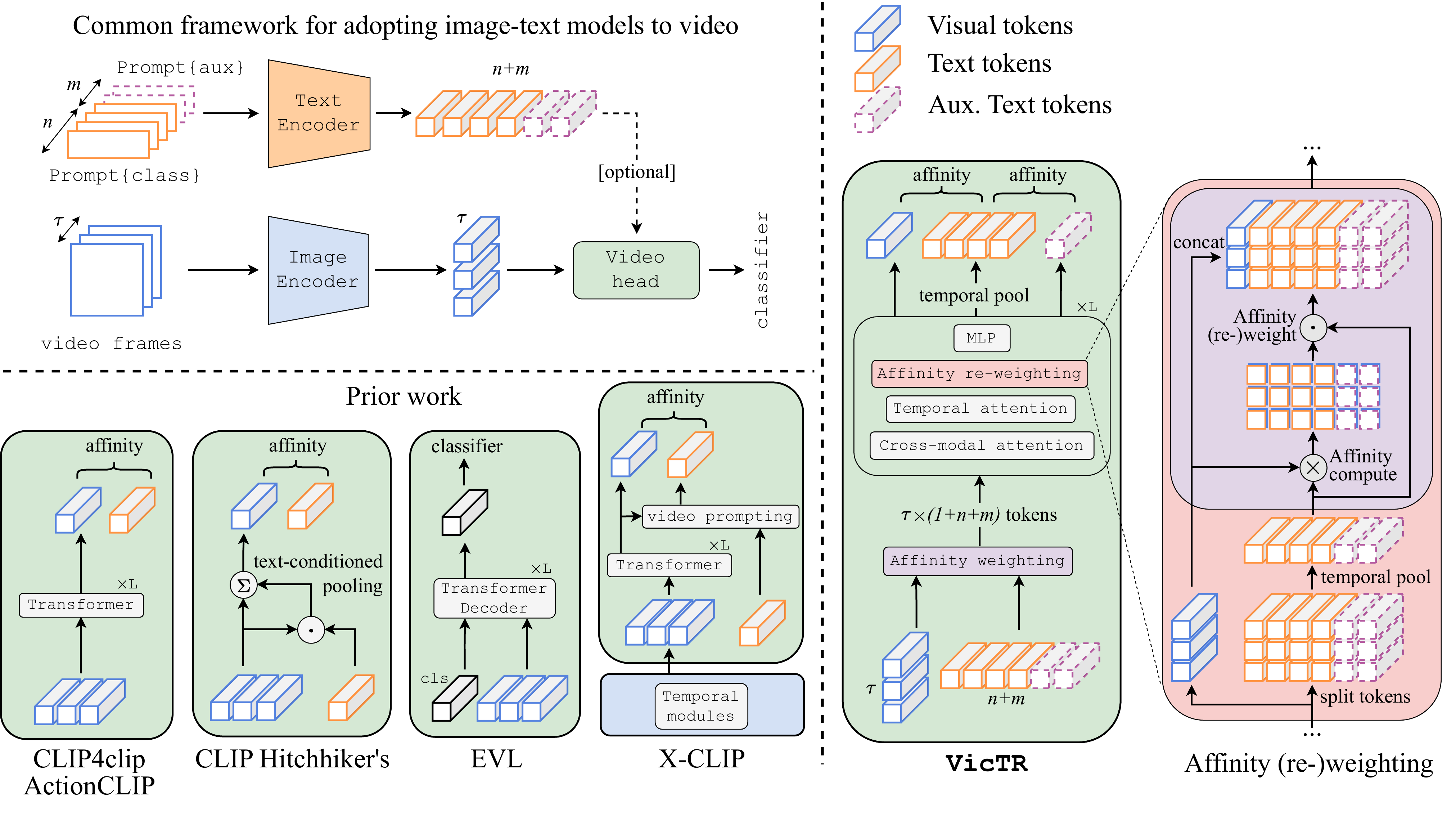}
    \vspace{-6mm}
    \caption{\textbf{Detailed view of \ours~compared to prior art:} There exist multiple closely-related work on adapting pretrained image-VLMs to video, such as CLIP4Clip \cite{luo2022clip4clip}, ActionCLIP \cite{wang2021actionclip}, CLIP Hitchhiker's \cite{bain2022cliphitchhiker}, EVL \cite{lin2022evl} and X-CLIP \cite{ma2022xclip}. All these follow a common framework (top-left). Text prompts and video frames are first encoded using two-separate encoders, and then fed into a video head to enable temporal reasoning. It is optional to use text tokens within the video head. Often, text information is kept unchanged \cite{luo2022clip4clip, wang2021actionclip}, or even discarded \cite{lin2022evl} (bottom-left). CLIP Hitchhiker's \cite{bain2022cliphitchhiker} however, use text as conditioning to generate \textit{text-conditioned video} embeddings. X-CLIP \cite{ma2022xclip}--- which is the closest to our method, jointly-optimizes visual and text tokens. But, it provides limited information for text to contrast-against: only temporally-aggregated visual embeddings, showing marginal gains from updating text. 
    In contrast, \ours~allows text to contrast against both fine-grained visual and other text information, while also jointly-optimizing both modalities. We generate \textit{video-conditioned text} representations, \ie text uniquely-specialized for each video (refer to \fref{fig:concept}). Our video head consists of three key operations: (1) \textit{Token-boosting}, (2) \textit{Cross-modal attention}, and (3) \textit{Affinity (re-)weighting}  (right). Token-boosting creates dedicated text tokens per video and per timestep, weighted by per-frame affinities of a given video. These enable us to model variations of semantics (represented as text) over time. Affinity (re)-weighting highlights or down-plays each text class, grounded on visual information. Such affinity weights are similar to the ones in CLIP \cite{radford2021clip} training objective, making the optimization more-consistent. Cross-modal attention enables message passing between both visual-textual and textual-textual modes, creating a better contrastive representation. Also, optionally, \ours~can make use of auxiliary semantics (\eg object, scene, human-subjects) given as visually-grounded text (refer to \fref{fig:overview}). Such auxiliary semantics help align our video-conditioned text embeddings in the latent space.
    }
    \label{fig:main}
    \vspace{-4mm}
\end{figure*}

In this section, we introduce the generic framework for adapting image-VLMs to video, and discuss how prior work fit into it. We consider CLIP \cite{radford2021clip} as the image-VLM, which is widely-adapted thanks to its convincing performance and open-source models. It consists of two encoders: Image and Text, optimized together on internet-scale paired image-text data. Image Encoder ($\texttt{Enc}_{\texttt{img}}$) is a ViT \cite{dosovitskiy2020vit}. Given an input image $I\in \mathbb{R}^{H\times W\times 3}$, it is broken down to patch embeddings (\ie tokens) and processed through multiple transformer layers. The class token $[\texttt{cls}]$ is sampled as the visual embedding $e_\texttt{img}$. Text Encoder ($\texttt{Enc}_{\texttt{txt}}$) is a causal transformer, operating on tokenized text. Each class-label (or, any semantic concept) given as text $T$, is first converted into a prompt based on a template such as ``\texttt{a photo of \{class\}.}'', and tokenized with Byte Pair Encoding (BPE) \cite{sennrich2015neural} at the input of Text Encoder. Following multiple causal transformer layers, the \texttt{[EOS]} (\ie end-of-sequence) token is extracted as the text embedding $e_\texttt{txt}$. %
\vspace{-2mm}
\begin{align*}
    e_\texttt{img} &= \texttt{Enc}_{\texttt{img}}(I), \\
    e_\texttt{txt} &= \texttt{Enc}_{\texttt{txt}}(T).
\end{align*}
The two encoders are jointly-optimized with Cross-Entropy loss, where logits are computed based on the similarities (\ie \textit{affinities}) between visual and text embeddings. The corresponding pairs of embeddings (\ie positives) are drawn together ($\uparrow$ affinity) in a joint embedding space, whereas the others (\ie negatives) are pushed apart ($\downarrow$ affinity).
\vspace{-2mm}
\begin{align*}
    \texttt{Affinity}(e_\texttt{img},\; e_\texttt{txt}) = \frac{\langle e_\texttt{img},\; e_\texttt{txt} \rangle}{\|e_\texttt{img}\|_2 \|e_\texttt{txt}\|_2}\;.
\end{align*}
When adapting this framework to the video domain, the above Image encoder, Text encoder and the learning objective usually stays the same. But now, video frames $V\in \mathbb{R}^{\Tau\times H\times W\times 3} = [I^1,\; I^2,\; \cdots,\; I^\Tau]$ become inputs to the Image encoder (while each being processed separately), and further go through a Video Head $\texttt{Head}_\texttt{vid}$ to induce temporal reasoning capabilities. Optionally, text embedding $e_\texttt{txt}$ may also be updated or used as a conditioning within the Video Head.
\begin{align*}
    e_\texttt{vid},\; [e_\texttt{txt}] &= \texttt{Head}_\texttt{vid}(e_{\texttt{img}}^1,\; \cdots,\; e_{\texttt{img}}^\Tau,\; [e_\texttt{txt}]).
\end{align*}
Here, $[\cdot]$ denotes optional embeddings. This Video Head may just be a temporal pooling layer or a temporal transformer as in \cite{luo2022clip4clip, wang2021actionclip}, or may even consist of more-specialized modules. Text embeddings could either be discarded as in \cite{lin2022evl}, used as a conditioning as in \cite{bain2022cliphitchhiker}, or jointly-updated with video embeddings as in \cite{ma2022xclip}. Finally, logits are computed based on video-text affinities if text is not discarded, or as a linear mapping of video embeddings if text is discarded. This generic framework is shown in \fref{fig:main} (top-left), along with variations of prior work in \fref{fig:main} (bottom-left).

\section{\textit{Video-conditioned Text} Representations}
\label{sec:method}

In \ours, we adapt a pretrained image-VLM (\eg CLIP \cite{radford2021clip}) to video, focusing more on text representations. Refer to \fref{fig:main} (right) for a detailed view. The image-VLM has not seen any temporal information during training. While it obviously affects the temporal reasoning capabilities of the visual embeddings--- which most prior work focus on addressing, it also affects the text embeddings as well. The learnt latent space (and, the affinity-based objective) depends on both these embeddings. Thus, we consider text equally as important, if not more, in contrast to prior work

\ours~consists of a joint video-text model as $\texttt{Head}_\texttt{vid}$, which consumes both visual and text embeddings from the image-VLM. It outputs text embeddings uniquely-specified for each video, \ie \textit{Video-conditioned Text} embeddings. It relies on three main components: (1) \textit{Token-boosting}, (2) \textit{Cross-modal attention}, and (3) \textit{Affinity (re-)weighting}. Optionally, it can also benefit from any semantic concept available as auxiliary text, to optimize its latent space. Following subsections look at each of these in detail.

Let us first introduce a few additional notations. Consider a fixed vocabulary of $n$ activity-classes given by $[T^1,\;T^2,\;\cdots,\;T^n]$, and optional $m$ auxiliary semantic categories given by $[A^1,\;A^2,\;\cdots,\;A^m]$. The corresponding text embeddings can be denoted as $\{e_\texttt{txt}^{x}\;|\;x=1, 2, \cdots, n\}$ and $\{e_\texttt{aux}^{y}\;|\;y=1, 2, \cdots, m\}$.
Also, given an input video $V^i$ of $\Tau$ frames, the corresponding image embeddings can be denoted as $\{e_\texttt{img}^{i,t}\;|\;t=1, 2, \cdots, \Tau\}$. The inputs to our Video Head are $e_\texttt{img}^{i,t}$, $e_\texttt{txt}^{x}$ and $e_\texttt{aux}^{y}$ tokens. As visual embeddings are extracted per-frame and the text embeddings per prompt, there is no interaction among frame tokens, among text tokens or, across frame-text tokens up to this point.

\subsection{Token-boosting}

To introduce \textit{video-conditioned text} embeddings, we first create a dedicated set of text tokens per video, by replicating the outputs of the backbone text encoder. Going further, we also create text tokens per each frame. This is done by weighting text tokens with the corresponding frame-text affinities. Formally, given $(n+m)$ text tokens, we end up with $\Tau\times(n+m)$ dedicated text tokens per video, at the input of our video head. Refer to \fref{fig:main} (right).
\begin{align*}
    e_\texttt{txt}^{i,t,x} = e_\texttt{txt}^{x} \cdot \texttt{SigAffinity}(e_\texttt{img}^{i,t},\; e_\texttt{txt}^{x}),\\
    e_\texttt{aux}^{i,t,y} = e_\texttt{aux}^{y} \cdot \texttt{SigAffinity}(e_\texttt{img}^{i,t},\; e_\texttt{aux}^{y}).%
\end{align*}
Here, \texttt{SigAffinity}($\cdot$) corresponds to affinity-weights normalized in $[0,1]$ range. We convert the values given by \texttt{Affinity}($\cdot$) that lie in $[-1,1]$, to be affinity-weights, by scaling with a learnable weight ($w$) and feeding through a sigmoid activation.
\begin{align*}
    \texttt{SigAffinity}(\cdot) = \texttt{Sigmoid}( w\cdot\texttt{Affinity}(\cdot)).
\end{align*}
Although such affinity-weights based on the original image-VLM embeddings are not ideal for temporal reasoning, it initializes a noisy-version of our \textit{video-conditioned text} embeddings that gets updated iteratively, later in the network. Such a token-boosting brings multiple other benefits. (1) More tokens means higher the model capacity. It can help learn better representations, but also adds a compute overhead (which we handle through other measures, as discussed later). (2) It also highlights relevant text tokens by grounding text on visual embeddings, while diminishing irrelevant ones. Subsequent attention mechanisms attend less to such diminished tokens, simplifying the gradient flow during learning. In other words, it acts as a soft-selection of relevant semantics, specific to each video. (3) Finally, it enables our model to capture variations of semantic categories over time. How certain attributes appear (or, disappear) over time is an important motion cue for activity recognition. 

Next, we concatenate such boosted text tokens with visual tokens (corresponding to $\Tau$ frames), and feed $\Tau\times(1+n+m)$ tokens to the subsequent layers.
\begin{align*}
    z^{i,t} = \texttt{Concat}&(e_\texttt{img}^{i,t},\; e_\texttt{txt}^{i,t,x}\big|_{x=\{1,\cdots,n\}},\; e_\texttt{aux}^{i,t,y}\big|_{y=\{1,\cdots,m\}}).
\end{align*}
Such $Z^i_0 = [z^{i,1},\; \cdots,\; z^{i,\Tau}]$ tokens go through $L$ transformer layers in our Video Head. Each layer ($l$) consists of cross-modal attention, temporal attention, affinity (re-)weighting and linear (MLP) layers.

\subsection{Cross-modal and Temporal attention}

We consider our token representation to be two-dimensional (\ie cross-modal and temporal), and apply divided self-attention (MSA) on each axis as in \cite{arnab2021vivit, bertasius2021timesformer}. First, we have a Cross-modal attention layer. Here, each visual token could attend to all text tokens at the same timestep, and each text token could attend to both the visual token and other text tokens at the same timestep. Since text tokens are already affinity-weighted, attention weights do not draw information from irrelevant semantic classes. Next, we have a Temporal attention layer. Here, both visual and text tokens go through a shared set of parameters, learning temporal cues in visual modality (\ie $e_\texttt{img} \rightarrow e_\texttt{vid}$), and modeling variations of semantics across time in textual modality. 
\begin{align*}
    \hat{Z}^i_l &= Z^i_l + \texttt{MSA}_\texttt{cross}(\texttt{LN}(Z^i_l)),\\
    \bar{Z}^i_l &= \hat{Z}^i_l + \texttt{MSA}_\texttt{temporal}(\texttt{LN}(\hat{Z}^i_l)).
\end{align*}
Here, $\texttt{LN}(\cdot$) stands for LayerNorm operation. Having a divided attention across two-axes instead of a joint-attention eases the compute requirement of our video head.

\subsection{Affinity (re-)weighting}

As previously discussed, the original affinities based on the image-VLM embeddings can be noisy, in the context of temporal reasoning. Now, as we have updated both our visual (\ie video) and text tokens with cross-modal and temporal information, they are in a better state to re-compute affinities. Hence, we compute new affinity values and re-weight the text tokens accordingly. Refer to \fref{fig:main} (rightmost). First, we split video and text tokens as in,
\begin{align*}
    \big[\bar{e}_{\texttt{vid},l}^{\;i,t},\; \bar{e}_{\texttt{txt},l}^{\;i,t,x}\big|_{x=\{1,\cdots,n\}}&,\; \bar{e}_{\texttt{aux},l}^{\;i,t,y}\big|_{y=\{1,\cdots,m\}}\big] = \bar{z}^{\;i,t}_l.
\end{align*}
Next, we temporally-pool the text tokens to come up with a compressed representation, on which we perform affinity re-weighting. This is similar to token-boosting, but done with updated video-text embeddings that are already video-conditioned. Without loss of generality, the same operations apply for auxiliary text tokens.
\begin{align*}
    \bar{e}_{\texttt{txt},l}^{\;i,x} &= \texttt{Pool}(\bar{e}_{\texttt{txt},l}^{\;i,t,x}),\\
    \bar{e}_{\texttt{txt},l}^{\;i,t,x} &= \bar{e}_{\texttt{txt},l}^{\;i,x} \cdot \texttt{SigAffinity}(\bar{e}_{\texttt{vid},l}^{\;i,t}, \;\bar{e}_{\texttt{txt},l}^{\;i,x}).
\end{align*}
Finally, such affinity (re-)weighted text tokens are concatenated with visual tokens, as $\bar{Z}^i_l$, and go through an MLP.
\vspace{-3mm}
\begin{align*}
     Z^i_{l+1} &= \bar{Z}^i_l + \texttt{MLP}(\bar{Z}^i_l).
\end{align*}

\subsection{Classifier}

Following $L$ transformer layers in our Video Head, we temporally-pool all tokens. We end up with a single video embedding, $n$ activity-text embeddings and $m$ aux-text embeddings. We further aggregate auxiliary embeddings, leaving a single embedding per each of the $k$ semantic categories (\eg object, scene, human-subjects). Finally, we compute logits based on affinity, similar to the CLIP \cite{radford2021clip} objective, and use Cross-Entropy loss for optimization. 
\begin{align*}
    \texttt{logit}^{i,x} &= \texttt{Affinity}(e_{\texttt{vid},L}^{i},\; e_{\texttt{txt},L}^{i,x}),\\
    \texttt{logit}_\texttt{aux}^{i,x,y} &= \texttt{Affinity}(e_{\texttt{txt},L}^{i,x},\; e_{\texttt{aux},L}^{i,y})\;\big|_{y=\{1,\cdots,k\}}.
\end{align*}

\subsection{Discussion on design decisions}

\paragraph{Auxiliary semantic information:} We rely on optional semantics (or, attributes) in the form of visually-grounded auxiliary text, to improve our \textit{video-conditioned text} embeddings. This is guided by the loss on $\texttt{logit}_\texttt{aux}$. The vocabulary of such auxiliary texts is fixed (\ie common for all videos) per dataset. On Charades, we consider 97 auxiliary text classes, and on Kinetics-400, we use 88 classes (refer the appendix for more details). To highlight only the relevant semantics for a given video, we visually-ground them via (1) cross-modal attention with visual embeddings, and (2) affinity weighting. Finally, to compute $\texttt{logit}_\texttt{aux}$, we create one \textit{representative embedding} per each of the $k$ semantic categories, by average pooling aux embeddings within a category ($k=4$ for Charades and $k=3$ for Kinetics-400).

\paragraph{Alternative weighting schemes:} Our text (re-)weighting method is similar to a contrastive training objective (as in CLIP \cite{radford2021clip}), which is based on visual-text affinities. We find this complementary nature beneficial. It highlights relevant text (and diminish irrelevant ones) within each intermediate layer of our Video Head. This iterative process fixes the initial noisy affinities resulting from the original image-VLM embeddings, when fused with better temporal cues in subsequent layers. We also explored other weighting schemes such as learnable weights or attention-based weights, which are not directly-connected to the training objective. They do not provide any improvements. 

\paragraph{Visual-only or Text-only classifiers:} We also explored different classifiers (\ie how we compute logits), considering (1) a visual-only classifier as in  \cite{lin2022evl}, (2) a text-only classifier, or (3) an affinity-based classifier as in \cite{radford2021clip, ma2022xclip}. The last performs the best. Even though we primarily focus on updating text embeddings, it still makes sense to rely on video-text affinities to be the training objective (or, classifier), as it is complementary to the components within our Video Head.

%% file: sections/5_experiments.tex
\section{Experiments}
\label{sec:experiments}

To validate the merits of \ours, we experiment on few-shot and zero-shot activity recognition (on HMDB-51 \cite{kuehne2011hmdb} and UCF-101 \cite{soomro2012ucf101}), as well as short-form (on Kinetics-400 \cite{kay2017kinetics}) and long-form recognition (on Charades \cite{sigurdsson2016hollywood}). Following sub-sections will detail our implementation, evaluation settings, datasets and the results.

\paragraph{Implementation details:} We use a pretrained CLIP \cite{radford2021clip} as our image-VLM backbone. Our Video Head is randomly-initialized having 4 transformer blocks similar to \cite{wang2021actionclip}, which is applied on-top of CLIP backbones. We consider an embedding dimension of 512/768 (w/ heads 8/12) corresponding to CLIP B/16 and L/14 backbone variants. Our output video-text embeddings are further mapped into 256-dimensional embeddings prior to computing affinity-based logits. We use an AdamW \cite{loshchilov2017adamw} optimizer with a cosine schedule for training. On Kinetics-400 \cite{kay2017kinetics}, we finetune our model for 30 epochs with a batch size of 256 using 8e-6/8e-5 learning rates for backbone/newly-initialized parameters, similar to \cite{ma2022xclip}. On Charades \cite{sigurdsson2016hollywood}, we finetune for 50k iterations with a batch size of 64 using 5e-7/5e-4 learning rates for backbone/newly-initialized parameters, similar to \cite{bain2022cliphitchhiker}. We use augmentations and input sampling strategies similar to \cite{ma2022xclip} for Kinetics-400 and similar to \cite{lin2022evl} for Charades.

\paragraph{Evaluation settings:} In our experiments, we compare against prior art VLMs on each dataset. Since the direction of adapting image-VLMs to video is relatively-recent, their absolute performance may not be the state-of-the-art in some cases (\eg long-form recognition), but we report numbers in comparable settings. For each experiment, we report pretraining settings, \#frames-per-view, \#views-at-inference and compute-per-view (GFLOPs) as supplementary metrics. We evaluate single-label activity recognition performance with Top-1 (\%) accuracy, and multi-label recognition with Average Precision (mAP\%). When reporting FLOPs, we consider the cost of computing a single affinty-based logit (\ie the cost for one video-text pair) similar to \cite{ma2022xclip}.

\subsection{Few-shot and Zero-shot Transfer}

\input{tables/few_shot}
\vspace{-1mm}
\paragraph{Data:} We consider the downstream datasets HMDB-51 \cite{kuehne2011hmdb} and UCF-101 \cite{soomro2012ucf101} to evaluate few-shot and zero-shot performance of our model. UCF-101 is a classification dataset collected from YouTube. It contains $\sim$13k clips annotated with 101 action classes. HMDB-51 is relatively small and contains $\sim$7k clips with 51 annotated classes. Both datasets have three splits of training/test data. In few-shot evaluation, we randomly sample 2, 4, 8, or 16 clips per class to create our training sets, same as in \cite{ma2022xclip}. We use a model pretrained on Kinetics-400 \cite{kay2017kinetics} for 10 epochs and finetune on few-shot examples for 50 epochs, using 32-frames per view as in \cite{ma2022xclip}. 

\vspace{-1mm}
\paragraph{Few-shot results:} In \tref{tab:fewshot}, we report top-1 accuracy on the first test split among three, in each dataset, using a single view at inference. \ours~significantly outperforms prior art, either w/o image-text pretraining (TSM \cite{lin2019tsm}, TimeSformer \cite{bertasius2021timesformer}, Video-Swin \cite{liu2022videoswin}) or w/ such pretraining (X-CLIP \cite{ma2022xclip}, X-Florence \cite{ma2022xclip}). Although our method uses similar backbones as X-CLIP, it even outperforms X-Florence (an extension of a more-generic foundation model) on both datasets consistently. This shows the effectiveness of our \textit{video-conditioned text} embeddings when generalizing to downstream with few training samples.

\input{tables/zero_shot}
\vspace{-1mm}
\paragraph{Zero-shot results:} We report zero-shot transfer performance in \tref{tab:zeroshot}. We use a model pretrained for 10 epochs on Kinetics-400 \cite{kay2017kinetics} with 32-frames per view, similar to  \cite{ma2022xclip}, and transfer to the downstream. We report mean and standard deviation on three-splits. \ours-B/16 outperforms X-CLIP \cite{ma2022xclip} by $6.4\%$ on HMDB-51 and by $0.4\%$ on UCF-101. Also, the performance of our model is more stable across splits. This validates that the learned \textit{video-conditioned} text embeddings can be generalized, even w/o seeing the same categories as in the downstream, during pretraining.

\subsection{Short-form Activity Recognition}

\input{tables/short_form}
\vspace{-1mm}
\paragraph{Data:} Kinetics-400 \cite{kay2017kinetics} is a large-scale activity recognition dataset, with 240k training and 20k validation videos. Each clip contains video-level annotations for a single activity out of 400 categories, having short $\sim$10s duration.

\vspace{-1mm}
\paragraph{Results:} We report the performance of \ours~on Kinetics-400 short-form activity recognition in \tref{tab:kin}. We consider L/14 with 8-frames per view, while using $4\times3$ such views at inference similar to \cite{ma2022xclip}. Our method shows a competitive performance at a similar footprint to closely-related video-VLMs \cite{ma2022xclip, lin2022evl}. 
It is also competitive with CoVeR-L \cite{zhang2021cover} which is trained with 10$\times$ more data. \ours~outperforms MTV \cite{yan2022mtv} by $+2.7\%$, ViViT \cite{arnab2021vivit} by $+3.5\%$ and TokenLearner \cite{ryoo2021tokenlearner} by $+1.6\%$, all trained on a similar scale of data, while being more-efficient.

\subsection{Long-form Activity Recognition}

\input{tables/long_form}
\vspace{-1mm}
\paragraph{Data:} Charades \cite{sigurdsson2016hollywood} is a small-yet-challenging activity recognition dataset with $\sim$9.8k long-form videos. It comes with frame-level annotations of 157 daily household activities. Yet, the benchmark setting requires making video-level predictions. The data is split as $\sim$7.9k for training and $\sim$1.8k for validation. Each video contains multiple overlapping activities, having an average duration of $\sim$30s.

\vspace{-1mm}
\paragraph{Results:} We report the performance of \ours~on Charades long-form activity recognition in \tref{tab:charades}. Here, we consider both B/16 and L/14 model variants with 32-frames per view, while having $4\times1$ such views at inference. Our method outperforms prior video-VLMs by a considerable margin. In fact, \ours-B/16 shows $+5.2\%$ mAP boost over CLIP Hitchhiker's \cite{bain2022cliphitchhiker}, and $+5.5\%$ mAP boost over ActionCLIP \cite{wang2021actionclip} with a similar footprint. This is a significant improvement considering the challenging Charades settings. Our method is also competitive with non-VLMs, whereas other video-VLMs lag behind. It highlights the limitations of current VLMs in long-context temporal modeling.

\subsection{Ablation Study}

\input{tables/ablate_main}
\input{tables/ablate_design}
In \tref{tab:ablations_main}, we provide evidence to validate our main hypotheses. Namely, we evaluate the impact of auxiliary semantics and the effectiveness of updating text embeddings.

\vspace{-1mm}
\paragraph{Auxiliary semantics do help.} We rely on extra semantic information to guide our latent embedding space. We see that such auxiliary text is giving $+0.2\%$ gain on Kinetics-400 and $+0.3\%$ mAP gain on Charades. This conveys the potential of semantics, but also the limitations of not having ground-truth annotations corresponding to them.

\vspace{-1mm}
\paragraph{Updating text embeddings is more effective.} To evaluate which of our embeddings (video or \textit{video-conditioned text}) are critical, we replace them with the corresponding original CLIP \cite{radford2021clip} embeddings (\ie temporally-pooled frame, or text). We see that the proposed \textit{video-conditioned text} are significantly-more effective, and when replaced, the performance drops $-1.1\%$ on Kinetics-400 and $-8.4\%$ mAP on Charades. In contrast, when our video embeddings are replaced, the performance drops only $-0.4\%$ and $-0.4\%$ mAP, respectively. Meaning, the CLIP frame embeddings are on-par with our video embeddings, but our \textit{video-conditioned text} embeddings are significantly improved.

\vspace{2mm}
In \tref{tab:ablations_sub}, we ablate and justify our design decisions. Namely, we evaluate our affinity weighting mechanism, divided attention, and affinity-based classifier.

\vspace{-1mm}
\paragraph{Affinity-weighting and divided attention do help.} We see a  $+1.3\%$ mAP performance gain by having our affinity (re-)weighting mechanism. While joint-attention may be more expressive compared to divided attention, it can incur training difficulties. As a result, we see the divided attention enjoying a significant $+5.3\%$ mAP boost.

\vspace{-1mm}
\paragraph{Affinity-based classifier is required.} As we previously discussed, our affinity weighting mechanism makes more-sense in the context of the same affinity-based loss formulation. To verify this, we replace such affinity-based logits with text-only or visual-only logits, which are just linear mappings of the corresponding embeddings. These significantly underperforms, with $-8.9\%$ mAP and $-7.0\%$ mAP, respectively.

%% file: tables/few_shot.tex
\newcommand\myeq{\mkern1.5mu{=}\mkern1.5mu}
\begin{table}[t!]
	\centering
	\tablestyle{1.8pt}{1.}
	\resizebox{1\linewidth}{!}{
		\begin{tabu}{l cccc c cccc}
		    \toprule
			\multirow{2.2}{*}{Model} & \multicolumn{4}{c}{HMDB-51} & & \multicolumn{4}{c}{UCF-101} \\
            \cmidrule(lr{1mm}){2-5}
            \cmidrule(l{1mm}){7-10}
            \multicolumn{1}{r}{$k$:} & $\;\;\;$2$\;\;\;$ & $\;\;\;$4$\;\;\;$ & $\;\;\;$8$\;\;\;$ & $\;\;$16$\;\;$ & & $\;\;\;$2$\;\;\;$ & $\;\;\;$4$\;\;\;$ & $\;\;\;$8$\;\;\;$ & $\;\;$16$\;\;$ \\
			\midrule
			\multicolumn{10}{l}{\textit{Methods w/o image-text pretraining}} \\
            \rowfont{\color{down}}TSM \cite{lin2019tsm} & 17.5 & 20.9 & 18.4 & 31.0 & & 25.3 & 47.0 & 64.4 & 61.0 \\
            \rowfont{\color{down}}TimeSformer \cite{bertasius2021timesformer} & 19.6 & 40.6 & 49.4 & 55.4 & & 48.5 & 75.6 & 83.7 & 89.4 \\
            \rowfont{\color{down}}Video-Swin-B \cite{liu2022videoswin} & 20.9 & 41.3 & 47.9 & 56.1 & & 53.3 & 74.1 & 85.8 & 88.7 \\
            \midrule
            \multicolumn{10}{l}{\textit{Methods w/ image-text pretraining}} \\
            X-CLIP \cite{ma2022xclip} & 53.0 & 57.3 & 62.8 & 64.0 & & 76.4 & 83.4 & 88.3 & 91.4 \\
            X-Florence \cite{ma2022xclip} & 51.6 & 57.8 & 64.1 & 64.2 & & 84.0 & 88.5 & 92.5 & 94.8 \\
            \rowcolor{row}\ours~(B/16) & 60.0 & 63.2 & 66.6 & 70.7 & & 87.7 & 92.3 & 93.6 & 95.8 \\
		    \bottomrule
	   \end{tabu}}
    \vspace{-2mm}
	\caption{\textbf{Few-shot Transfer:} On HMDB-51 \cite{kuehne2011hmdb} and UCF-101 \cite{soomro2012ucf101}, we compare our method against prior art, reporting top-1 accuracy (on the first split among three test splits as in \cite{ma2022xclip}). We use models pretrained on Kinetics-400 \cite{kay2017kinetics} for 10 epochs, and finetune on few-shot samples for 50 epochs. We randomly-sample $k=\{2,\;4,\;8,\;16\}$ clips per class as few-shot training samples in each setting. \ours~shows a significant boost over X-CLIP \cite{ma2022xclip}. Non-VLMs are \textcolor{down}{de-emphasized}.}
    \label{tab:fewshot}
    \vspace{3mm}
\end{table}

%% file: tables/zero_shot.tex
\begin{table}[t!]
	\centering
	\tablestyle{1.8pt}{1.}
	\resizebox{0.65\linewidth}{!}{
		\begin{tabu}{lccc}
		    \toprule
			Model & \#Frames & HMDB-51 & UCF-101 \\
			\midrule
			\multicolumn{4}{l}{\textit{Methods w/o image-text pretraining}} \\
			\rowfont{\color{down}}MTE \cite{xu2016mte} & - & 19.7{\scriptsize $\;\pm\;$1.6} & 15.8{\scriptsize $\;\pm\;$1.3} \\
			\rowfont{\color{down}}ASR \cite{wang2017asr} & 16 & 21.8{\scriptsize $\;\pm\;$0.9} & 24.4{\scriptsize $\;\pm\;$1.0} \\
            \rowfont{\color{down}}ZSECOC \cite{qin2017zsecoc} & - & 22.6{\scriptsize $\;\pm\;$1.2} & 15.1{\scriptsize $\;\pm\;$1.7} \\
            \rowfont{\color{down}}UR \cite{zhu2018ur} & 1 & 24.4{\scriptsize $\;\pm\;$1.6} & 17.5{\scriptsize $\;\pm\;$1.6} \\
            \rowfont{\color{down}}TS-GCN \cite{gao2019tsgcn} & 16 & 23.2{\scriptsize $\;\pm\;$3.0} & 34.2{\scriptsize $\;\pm\;$3.1} \\
            \rowfont{\color{down}}E2E \cite{brattoli2020e2e} & 16 & 32.7$\;\;\;\;\;\;\;\:\:$ & 48.0$\;\;\;\;\;\;\;\;\:$ \\
            \rowfont{\color{down}}ER-ZSAR \cite{chen2021erzsar} & - & 35.3{\scriptsize $\;\pm\;$4.6} & 51.8{\scriptsize $\;\pm\;$2.9} \\
            \midrule
		    \multicolumn{4}{l}{\textit{Methods w/ image-text pretraining}} \\
            ActionCLIP \cite{wang2021actionclip} & 32 & 40.8{\scriptsize $\;\pm\;$5.4} & 58.3{\scriptsize $\;\pm\;$3.4} \\
            X-CLIP \cite{ma2022xclip} & 32 & 44.6{\scriptsize $\;\pm\;$5.2} & 72.0{\scriptsize $\;\pm\;$2.3} \\
            \rowcolor{row}\ours~(B/16) & 32 & 51.0{\scriptsize $\;\pm\;$1.3} & 72.4{\scriptsize $\;\pm\;$0.3} \\
		    \bottomrule
	\end{tabu}}
    \vspace{-2mm}
	\caption{\textbf{Zero-shot Transfer:} On HMDB-51 \cite{kuehne2011hmdb} and UCF-101 \cite{soomro2012ucf101}, we compare our method against prior art, reporting input format (\#Frames) and top-1 accuracy (\%) as mean$\pm$std across the three splits of test set as in \cite{ma2022xclip}. We use models pretrained on Kinetics-400 for 10 epochs. \ours~outperforms similar video-VLM adaptations. Non-VLMs are \textcolor{down}{de-emphasized}.}
    \label{tab:zeroshot}
    \vspace{-1mm}
\end{table}

%% file: tables/short_form.tex
\begin{table}[t!]
	\centering
	\tablestyle{1.8pt}{1.}
	\resizebox{0.98\linewidth}{!}{
		\begin{tabu}{lccccr}
		    \toprule
			\multicolumn{1}{l}{Model}  & Pretrain & \#Frames & \#Views & GFLOPs & Top-1 \\
			\midrule
			\multicolumn{6}{l}{\textit{Methods w/o image-text pretraining}} \\
			\rowfont{\color{down}} Video-Swin-L (384$\uparrow$) \cite{liu2022videoswin} & IN-21K & 32 & 10$\times$5 & 2107 & 84.9 \\
            \rowfont{\color{down}} TimeSformer-L \cite{bertasius2021timesformer} & IN-21K & 96 & 1$\times$3 & 2380 & 80.7 \\
			\rowfont{\color{down}} MTV-L \cite{yan2022mtv} & JFT-300M & 32 & 4$\times$3 & 1504 & 84.3 \\
			\rowfont{\color{down}} Video-SwinV2-G (384$\uparrow$) \cite{liu2022videoswin} & IN-21K+ & 8 & 4$\times$5  & - & 86.8 \\
			\rowfont{\color{down}} MViTv2-L \cite{li2022mvitv2} (312$\uparrow$) & - & 40 & 5$\times$3 & 2828 & 86.1 \\
   			\rowfont{\color{down}} ViViT-L FE \cite{arnab2021vivit} & JFT-300M & 32 & 1$\times$3 & 3980 & 83.5 \\
            \rowfont{\color{down}} TokenLearner \cite{ryoo2021tokenlearner} & JFT-300M & 64 & 4$\times$3 & 4076 & 85.4 \\
			\rowfont{\color{down}} CoVeR-L \cite{zhang2021cover} & JFT-3B & - & 1$\times$3 & - & 87.2 \\
			\midrule
			\multicolumn{6}{l}{\textit{Methods w/ image-text pretraining}} \\
			ST-Adapter \cite{pan2022stadapter} & CLIP & 32 & 1$\times$3 & 2749 & 87.2 \\
			Text4Vis \cite{wu2022text4vis} & CLIP & 32 & 1$\times$3 & 1662 & 87.1 \\
			EVL \cite{lin2022evl} & CLIP & 8 & 1$\times$3 & 674 & 86.3 \\
			X-CLIP \cite{ma2022xclip} & CLIP & 8 & 4$\times$3 & 658 & 87.1 \\
			\rowcolor{row}\ours~(L/14) & CLIP & 8 & 4$\times$3 & 656 & 87.0  \\
			\bottomrule
	\end{tabu}}
    \vspace{-1mm}
	\caption{\textbf{Short-form Activity Recognition:} On Kinetics-400 \cite{kay2017kinetics}, we compare our method against prior art, reporting pretraining settings, input format, compute cost (GFLOPs) and top-1 accuracy (\%). Here, \#Frames represents the number of frames per view, while \#Views represents the number of temporal$\times$spatial crops during inference. The compute cost is reported per view. \ours~ shows a competitive performance among video-VLMs with a similar cost. Non-VLMs are \textcolor{down}{de-emphasized}.}
    \label{tab:kin}
\end{table}

%% file: tables/long_form.tex
\begin{table}[t!]
	\centering
	\tablestyle{1.8pt}{1.}
	\resizebox{0.95\linewidth}{!}{
		\begin{tabu}{lccccr}
		    \toprule
			\multicolumn{1}{l}{Model}  & Pretrain & \#Frames & \#Views & GFLOPs & mAP \\
			\midrule
			\multicolumn{6}{l}{\textit{Methods w/o image-text pretraining}} \\
			\rowfont{\color{down}} I3D + NL \cite{wang2018nonlocal} & K400 & 128 & 10$\times$3 & 544 & 37.5 \\
			\rowfont{\color{down}}EvaNet \cite{piergiovanni2019evanet} & K400 & 64 & - & - & 38.1 \\
			\rowfont{\color{down}}LFB-101 \cite{wu2019lfb} & K400 & 32 & 10$\times$3 & 529 & 42.5 \\
			\rowfont{\color{down}}SlowFast-50 \cite{feichtenhofer2019slowfast} & K400 & 8+32 & 10$\times$3 & 66 & 38.0 \\
			\rowfont{\color{down}}SlowFast-101 + NL \cite{feichtenhofer2019slowfast} & K400 & 16+64 & 10$\times$3 & 234 & 42.5 \\
			\rowfont{\color{down}}X3D-XL (312$\uparrow$) \cite{feichtenhofer2020x3d} & K400 & 16 & 10$\times$3 & 48 & 43.4 \\
			\rowfont{\color{down}}MViT \cite{fan2021mvit} & K400 & 32 & 10$\times$3 & 237 & 47.7 \\
		    \rowfont{\color{down}}AssembleNet-101 \cite{ryoo2019assemblenet} & - & 128 & 5$\times$1 & 1200 & 58.6 \\
			\midrule
			\multicolumn{6}{l}{\textit{Methods w/ image-text pretraining}} \\
			ActionCLIP \cite{wang2021actionclip} & CLIP & 32 & 10$\times$3 & 563 & 44.6 \\
			CLIP4clip \cite{luo2022clip4clip} & CLIP & 32 & 1$\times$1 & - & 32.0 \\
			CLIP Hitchhiker's \cite{bain2022cliphitchhiker} & CLIP & 32 & 1$\times$1 & - & 44.9 \\
			\rowcolor{row}\ours~(B/16) & CLIP & 32 & 4$\times$1 & 567 & 50.1 \\
			\rowcolor{row}\ours~(L/14) & CLIP & 32 & 4$\times$1 & 2602 & 57.6 \\
			\bottomrule
	\end{tabu}}
    \vspace{-1mm}
	\caption{\textbf{Long-form Activity Recognition:} On Charades \cite{sigurdsson2016hollywood}, we compare our method against prior art, reporting pretraining settings, input format, compute cost (GFLOPs) and mean Average Precision (mAP\%). The compute cost reported is per view. Here, \#Views represents the number of temporal$\times$spatial crops, each having \#Frames per view). \ours~ achieves strong a performance among the methods pretrained w/ image-text data by a considerable margin. Non-VLMs are \textcolor{down}{de-emphasized}.}
	\label{tab:charades}
    \vspace{2mm}
\end{table}

%% file: tables/ablate_main.tex
\begin{table}[t!]
	\centering
	\tablestyle{1.8pt}{1.}
	\resizebox{0.75\linewidth}{!}{
	    \tablestyle{1.8pt}{1.}
        \fontsize{8.8}{11}\selectfont
		\begin{tabular}{lcc}
		\toprule
		Model & Kinetics-400 & Charades \\
		\midrule
		\rowcolor{row}\ours & 84.4 & 50.1 \\
		\ours~ (No Aux. Text) & 84.2 & 49.8 \\
		\ours~ (w/ CLIP Visual emb.) & 84.0 & 49.7 \\
		\ours~ (w/ CLIP Text emb.) & 83.3 & 41.7 \\
		\bottomrule
        \end{tabular}
        }
    \vspace{-1mm}
	\caption{\textbf{Ablating main hypotheses:} On Kinetics-400 \cite{kay2017kinetics} and Charades \cite{sigurdsson2016hollywood}, we measure the importance of auxiliary text prompts. We also show that updating text is most critical in our framework, rather than updating visual embeddings (i.e., temporally-pooled CLIP image embeddings is as good as our video embeddings).}
    \vspace{-2mm}
    \label{tab:ablations_main}
\end{table}

%% file: tables/ablate_design.tex
\begin{table}[t!]
	\centering
	\tablestyle{1.8pt}{1.}
	\resizebox{0.54\linewidth}{!}{
	    \tablestyle{1.8pt}{1.}
        \fontsize{8.8}{11}\selectfont
		\begin{tabular}{lr}
		\toprule
		Model & mAP \\
		\midrule
		\rowcolor{row}\ours & 50.1 \\ %
		\ours~ (No Affinity weighting) & 48.8 \\
		\ours~ (w/ joint-attention) & 44.8 \\ 
		\ours~ (Text Classifier) & 41.2 \\
		\ours~ (Visual Classifier) & 43.1 \\ 
		\bottomrule
        \end{tabular}
        }
    \vspace{-1mm}
	\caption{\textbf{Ablating design decisions:} On Charades \cite{sigurdsson2016hollywood}, we evaluate different design decisions of \ours. First, we show the effectiveness of Affinity Weighting and divided attention in our framework. We also replace our visual-text affinity-based logits with simpler visual-only or text-only logits to show the benefit of ours.}
    \vspace{-3mm}
    \label{tab:ablations_sub}
\end{table}

%% file: sections/6_conclusion.tex
\section{Conclusion}
\label{sec:conclusion}

In this paper, we introduced \ours, a framework for adapting image-VLMs to video, with a focus on \textit{video-conditioned text} embeddings. It can also benefit from freely-available auxiliary semantic information in the form of visually-grounded text, to guide the learned latent space. Our evaluations verified the importance of updating text embeddings, across multiple activity recognition benchmarks, under few-shot, zero-shot, short-form and long-form settings. We believe that this work reveals the importance of using language embeddings for temporal reasoning.

%% file: sections/7_appendix.tex
\setcounter{table}{0}
\setcounter{figure}{0}
\setcounter{section}{0}
\renewcommand{\thetable}{A.\arabic{table}}
\renewcommand{\thefigure}{A.\arabic{figure}}
\renewcommand{\thesection}{A}
\renewcommand{\thesubsection}{A.\arabic{subsection}}

\section*{Appendix}

\paragraph{Details on auxiliary text classes:} On Charades \cite{sigurdsson2016hollywood}, we use 97 auxiliary classes: 43 objects, 15 places, 5 people-counts and 34 atomic-actions. People-count prompts are manually-selected, whereas the others are already annotated in the dataset. On Kinetics-400 \cite{kay2017kinetics}, we use 88 auxiliary classes: 40 objects, 43 places and 5 people-counts. Atomic-actions on Kinetics-400 are too diverse to be categorized as a concise set, and thus omitted. On Kinetics-400, people-counts are similarly selected, and the others are generated by prompting ChatGPT3.5 with the set of 400 activity classes. The auxiliary vocabulary for each dataset is given below.

\vspace{2mm}
\noindent On Charades \cite{sigurdsson2016hollywood}, we have the following: 

\noindent \underline{\texttt{Objects}}: \textit{bag, bed, blanket, book, box, broom, chair, closet, cabinet, clothes, cup, glass, bottle, dish, door, doorknob, doorway, floor, food, groceries, hair, hands, laptop, light, medicine, mirror, paper, notebook, phone, camera, picture, pillow, refrigerator, sandwich, shelf, shoe, sofa, couch, table, television, towel, vacuum, window.} 

\noindent \underline{\texttt{Places}}:  \textit{basement, garage, pantry, recreation room, walk-in closet, laundry room, stairs, hallway, dining room, entryway, home office, bathroom, kitchen, bedroom, living room.} 

\noindent \underline{\texttt{People}}: \textit{no people, one person, two people, three people, several people.} 

\noindent \underline{\texttt{Atomic-actions}}: \textit{doing nothing, awakening, closing, cooking, dressing, drinking, eating, fixing, grasping, holding, laughing, lying, making, opening, photographing, playing, pouring, putting, running, sitting, smiling, sneezing, snuggling, standing, taking, talking, throwing, tidying, turning, undressing, walking, washing, watching, working.}

\vspace{2mm}
\noindent On Kinetics-400 \cite{kay2017kinetics}, we have the following:

\noindent \underline{\texttt{Objects}}: \textit{bow and arrow, flowers, leaves or tree, computer, bed or baby crib, glass or bottle, dumbbell, treadmill or gym equipment, trampoline, mechanical bull or roller skates, bowling ball, cabinet or windows or dining table, sailboat or jet ski, fishing rod, cleaning supplies, grooming tools, pool, shoes, toilet, rope or ladder, barbecue grill or campfire, makeup tools, shovel, laundry or clothes, books or drawing materials, baseball, basketball or golf club, gymnastics mat, ice skates, dessert, fruits or vegetables, food items, fire extinguisher, hammer or meat grinder, musical instruments, board game, sporting equipment, gas pump, shopping cart, newspaper, animals, car, tractor or bicycle, rock climbing gear, electric sharpener or shredder.}

\noindent \underline{\texttt{Places}}: \textit{home, living room, dining room, bathroom, kitchen, bedroom, backyard or garden, staircase, hair salon, restaurant, outdoor, mountain or cliff, grass field, snow or ice, river or sea, sky, gym or fitness center, supermarket, foundary or workshop, forest, sports field, stadium, court or arena, massage palor, dance floor or stage, road or sidewalk, swimming pool, restaurant or bar, entrance or doorway, hospital or emergency room, bowling alley, building or skyscraper, theatre or auditorium, farm, recording studio or music room, news room, repair shop, garage, archery or shooting range, beach, underwater or sea bed, office or workspace, park, arcade or casino, school or classroom.}

\noindent \underline{\texttt{People}}: \textit{no people, one person, two people, three people, several people.}

\paragraph{On the selection of datasets:}
In literature, activity recognition is considered as the prominent video classification task. To understand the effectiveness of our \textit{video-conditioned text} representations, we tackle a variety of activity recognition benchmarks. This includes few-shot and zero-shot activity recognition (on HMDB-51 \cite{kuehne2011hmdb}, UCF-101 \cite{soomro2012ucf101}), short-form recognition (on Kinetics-400 \cite{kay2017kinetics}) and long-form recognition (on Charades \cite{sigurdsson2016hollywood}). It is worth noting that Kinetics-400 usually contains single-person activities, whereas Charades includes multiple people and complex overlapping activities. Together, these provide a thorough spread of scenarios for both single-label and multi-label classification. Our evaluation setting is similar to many other prior work which evaluate on classification \cite{wang2021actionclip, ma2022xclip, lin2022evl}, yet extensive as it includes diverse contexts.

\paragraph{Compute requirement:} Token-boosting increases the footprint of our model. However, our Video-Head is still lightweight, requiring minimal additional computations. In fact, it amounts for only 0.2\% (0.5B) of total FLOPs in B/16 16-frame model (285B), and only 0.1\% (0.6B) in L/14 8-frame model (656B). This is because of three reasons: (1) having fewer layers (\ie 4 layers vs.~12/24 layers) and lightweight attention modules (\ie temporal and cross-modal attention vs.~spatial attention) compared to the image-VLM backbone \cite{radford2021clip}, (2) processing significantly fewer tokens (\ie only temporal and text-class tokens remain), and (3) doing text-conditioning only after the backbone (\ie for the most part, all text embeddings go through shared computations). Ovrall, \ours~has a comparable footprint to prior work such as \cite{lin2022evl, ma2022xclip, wang2021actionclip}, providing a fair comparison (see respective GFLOPs in \tref{tab:kin} and \tref{tab:charades}).

\paragraph{Other forms of semantic information:} In our framework, we use a fixed vocabulary of auxiliary prompts as semantic inputs, that is specific to each dataset. Another way of providing semantic information is in the form of captions. If available, a detailed set of captions may provide better semantic supervision. However, they come with a significant cost, since they need to be annotated per-video. In contrast, our auxiliary prompts are freely-available and can be selected with only a minimal effort, as they are common for all videos in a dataset. Our model learns to highlight relevant information for a given video implicitly, via affinity weighting, without needing any ground-truth annotations.

\input{tables/rich_text}
\paragraph{Impact of more-descriptive text:} By default, we use class labels with the standard CLIP \cite{radford2021clip} prompt template to generate text embeddings. However, if available, more-descriptive text such as human-annotated captions (expensive) or machine-generated descriptions (inexpensive) can provide richer information for our cross-modal attention, improving \textit{video-conditioned text} representations. We validate this claim by replacing class-labels with rich class-descriptions from ChatGPT3.5 (\tref{tab:rich_text}). On zero-shot evaluation, the relative gains from our text improve on both HMDB-51 \cite{kuehne2011hmdb} (+7.1\% $\rightarrow$ +8.2\%) and  UCF-101 \cite{soomro2012ucf101} (+5.2\% $\rightarrow$ +6.7\%), also raising the absolute performance.

\input{tables/vqa}
\paragraph{Other reasoning tasks:} 
The primary scope of this paper is on a broad spectrum of recognition tasks. Yet, it is also applicable to other reasoning tasks such as video VQA. In \tref{tab:vqa}, we evaluate \ours~on NExT-QA \cite{xiao2021nextqa} under zero-shot settings, showing gains over comparable baselines with encoder-only designs (\ie no LLM decoders). This validates that our model can readily be extended to other tasks with jointly-embedded video and text.

%% file: tables/rich_text.tex
\begin{table}[t!]
	\centering
	\tablestyle{1.8pt}{1.}
	\resizebox{0.87\linewidth}{!}{
		\begin{tabu}{lccc}
		    \toprule
            \multicolumn{1}{l}{Model} & Rich text & $\;$HMDB-51$\;$ & $\;$UCF-101$\;$ \\
            \midrule
            X-CLIP \cite{ma2022xclip} & \xmark & 44.6{\scriptsize $\;\pm\;$5.2} & 72.0{\scriptsize $\;\pm\;$2.3} \\
            \ours~(w/ CLIP Text emb.) & \xmark & 43.9{\scriptsize $\;\pm\;$0.7} & 67.2{\scriptsize $\;\pm\;$0.7} \\
            \rowcolor{row}\ours & \xmark & 51.0{\scriptsize $\;\pm\;$1.3} & 72.4{\scriptsize $\;\pm\;$0.3} \\
            \midrule
            \ours~(w/ CLIP Text emb.) & \cmark & 43.9{\scriptsize $\;\pm\;$1.5} & 70.7{\scriptsize $\;\pm\;$0.3} \\
            \rowcolor{row}\ours & \cmark & 52.1{\scriptsize $\;\pm\;$0.5} & 77.4{\scriptsize $\;\pm\;$0.2} \\ 
			\bottomrule
	\end{tabu}}
    \vspace{-2mm}
	\caption{\textbf{Impact of more-descriptive text:} We replace class labels in HMDB-51 \cite{kuehne2011hmdb} and UCF-101 \cite{soomro2012ucf101} with rich class-descriptions generated by ChatGPT3.5. On zero-shot evaluation, our video-conditioned text embeddings benefit significantly-more from rich text inputs, compared to the CLIP \cite{radford2021clip} text embeddings.}
    \vspace{2mm}
    \label{tab:rich_text}
\end{table}

%% file: tables/vqa.tex
\begin{table}[t!]
	\centering
	\tablestyle{1.8pt}{1.}
	\resizebox{0.64\linewidth}{!}{
		\begin{tabu}{lccc}
		    \toprule
            \multicolumn{1}{l}{Model} & Type & Params & NExT-QA \\
            \midrule
            Random & - & - & 20.0 \\
            \midrule 
            \rowfont{\color{down}}CaKE-LM \cite{Su2023cakelm} & \multirow{3}{*}{Enc-Dec} & 2.7B & 34.9 \\
            \rowfont{\color{down}}InternVideo \cite{internvideo} & & 1.3B & 49.1 \\
            \rowfont{\color{down}}SeViLA \cite{yu2024sevila} & & 4.1B & 63.6 \\
            \midrule 
            Just-Ask \cite{yang2021justask} & \multirow{3}{*}{Enc only} & 75M & 38.4 \\
            X-CLIP \cite{ma2022xclip} & & 194M & 43.8  \\
            \rowcolor{row}\ours~(B/16) & & 167M & 45.5 \\
			\bottomrule
	\end{tabu}}
    \vspace{-2mm}
	\caption{\textbf{Video reasoning with VQA:} On NExT-QA \cite{xiao2021nextqa} zero-shot evaluation, our model outperforms comparable baselines. Large-scale models with LLM decoders are \textcolor{down}{de-emphasized}.}
    \vspace{-1mm}
    \label{tab:vqa}
\end{table}